# An open GPS trajectory dataset and benchmark for travel mode detection

Jinyu Chen[1], Haoran Zhang[1], Xuan Song[2, 1], Ryosuke Shibasaki[1]

1. Center for Spatial Information Science, The University of Tokyo, 5-1-5 Kashiwanoha, Kashiwa, Chiba 277-8563, Japan
2. SUSTech-UTokyo Joint Research Center on Super Smart City, Department of Computer Science and Engineering, Southern University of Science and Technology (SUSTech), Shenzhen, China

**Abstract**: Travel mode detection has been a hot topic in the field of GPS trajectory-related processing. Former scholars have developed many mathematical methods to improve the accuracy of detection. Among these studies, almost all of the methods require ground truth dataset for training. A large amount of the studies choose to collect the GPS trajectory dataset for training by their customized ways. Currently, there is no open GPS dataset marked with travel mode. If there exists one, it will not only save a lot of efforts in model developing, but also help compare the performance of models. In this study, we propose and open GPS trajectory dataset marked with travel mode and benchmark for the travel mode detection. The dataset is collected by 7 independent volunteers in Japan and covers the time period of a complete month. The travel mode ranges from walking to railway. A part of routines are traveled repeatedly in different time slots to experience different road and travel conditions. We also provide a case study to distinguish the walking and bike trips in a massive GPS trajectory dataset.

## 1 Introduction

Travel mode detection in GPS trajectory is long term hot topic[1, 2]. Stenneth et al.[3] developed a hybrid Bayes network merged with decision tree to distinguish the travel mode. Widhalm et al.[4] created a decision trees followed by a Discrete Hidden Markov Model and proposed 77 features for distinguishment. Zong et al.[5] used nested logit model and its variances. former studies have developed enormous methodologies to detect travel mode. A successful detection on travel mode not only requires reliable method, but also a comprehensive dataset that covers various situation. In most previous works[6-8], authors operated their own custom dataset collection and train the model. However, currently, there is barely open dataset for training of detection model. Because of various dataset, it's hard to preciously compare the performance of different detection model. In this work, we provide and introduce an open GPS trajectory dataset marked with travel mode as the training dataset for the travel mode detection. The dataset is collected by 7 independent volunteers and covers the time period of a complete month. The travel mode ranges from walking to railway.

In this work, we adopted the classical random forest method to distinguish the bike travels in non-motorized trips. The whole process consists of ground truth data collection, velocity analysis and training of detection model.

## 2 Ground truth data collection

A considerable part of works in travel mode detection employed the collection of GPS data of different travel modes by using different kinds of devices like smartphone[9], GPS logger[1]. Therefore, we also operated the ground truth collection of GPS data considering both bike travelling and walking by using Android smartphone. The accuracy of GPS record is $10^{-7}$ m in open areas. This refers to about 2-3 meters on surface of earth, which is enough for study. The time interval of GPS record is set to be 1 second.

The data collection covers the whole month of October 2020. There are totally 212 walking trips, 138 bicycle trips, 56 bus trips and 69 railway trips. The least durance of each trip is required to be 10 minutes.

Firstly, we introduce the bike trips. The bike trips are mainly located in a classical urban and rural fringe. There are both downtown which contains high density of buildings and high traffic volume as well as remote rural areas where there are residential houses, industrial factories and campuses. The distribution of bike trips is shown in Figure 5. A part of roads are repeatedly travelled by various times at different time slots to experience different road condition.

Figure 1. The trips of bike in the ground truth data

The walking trips are scatteredly distributed in places that are away from each other. Therefore, we choose to mainly show the representative samples of trips in urban and rural fringe and urban area in Figure 6.

Figure 2. Samples of walking trips (a) Urban and rural fringe (b) Urban area

The part of urban and rural fringe is similar with bike trips. In urban area, the walking trips cover a majority part of places of interest in the greater Tokyo area. The places marked with purple are the main parks, where residents and tourists usually visit for leisure and tourism; The orange ones are the major sub business district areas. Because of great traffic volumes and high accessibility, walking is also a very common travel behavior in these areas. Other walking trips are located in some corners like the Hongo Campus of the University of Tokyo.

Railway trips are also distributed both in rural and crowded areas. The spatial distribution of railway trips is shown in Figure 3. Trips with shape of blue line are on the railways that connect different urban centers or subcenters. For example, Joban Line connects the city center of Kashiwa and sub center of Tokyo; Tobu Urban Park Line and Keiyo Line connect the city center of Kashiwa and center. Trips with shape of green line are one the railways that are all in the crowded urban area. Difference is that the distance among stations on center-center lines is larger, so that the travel time among lines is shorter. The train usually needs to go through rural areas where there is no station. However, the distance on urban lines is the opposite. Therefore, the feature of travelling on two lines including average distance, travel time are different. In the groundtruth dataset, we include trips on both kinds of lines. Some lines are repeatedly traveled in different time slots.

Figure 3. Spatial distribution of railway trips

Finally, we introduce the trajectory dataset of bus trips. The trajectories mainly cover Tokyo, a metropolis, and Kashiwa city, a normal and average level city in Chiba prefecture, Japan. The trajectories with blue line rectangle are on the bus lines that connect downtown and suburbs The ones with green line rectangle are on the bus lines that operate fully in downtown and crowded commercial areas. Like railway, the travel pattern is also different between two kinds of bus lines.

Figure 3. Spatial distribution of bus trips

In summary, the ground truth trips cover various kinds of urban districts with different traffic volumes. They range from crowded sub business districts in city center to remote industrial and residential areas. The collection is suitable for the analysis of real travel behaviors.

## 3  Case study

In this study, we use our dataset to train a model for travel mode detection and try to distinguish the bike and walking trips in the massive GPS trajectory dataset.

**Velocity analysis**

The raw ground truth data collection always needs a preprocess. Because of the bad signal of GPS device in the smartphone, sometimes the error of recorded position could be too enormous. In the raw data, the estimated error of each recorded position is also recorded by the application in the smart phone as an independent column. We operated statistics on the recorded errors shown as Figure 7.

Figure 7. Boxplot of recorded errors

Under normal circumstances, the error isn't too large. The Q3 value is lesser than 50 m. On the other hand, the upper bound of error is computed to be 93.1 m. Values higher than 93.1 probably refer to the ones with abnormal errors in position. Therefore, in the preprocess, we removed these records.

Then, to fit the ground truth trip collection to the DOCOMO trajectory dataset, we subsample the ground truth trajectory by the interval of multiple of 1 minute until 5 minutes. The logic of subsampling is:

1. Set time interval of subsampling
2. Set the first GPS record as initiation of sampled trajectory and reference point
3. Sample the point with time interval or point closest to it if it does not exist in the original trajectory for sampling.
4. Set the sampled point as new reference point.

Then, in the subsampled trajectory, we compute the shortest distance between the neighboring points and divide it by their time interval as the average velocity from the previous one to latter one. Here, we show illustrate the change of distribution of velocity in walking and bike trips due to the increase of time interval.

Figure 8. Distribution of velocity under different time interval

It is easy to obvious from figure that with the increase of time interval, the distribution of velocity of walking and bike trips are getting closer. The KS test statistic value of both distributions decrease from 0.732 to 0.597 as the time interval increases from 1 minute to 5 minutes. So, there is an obvious approaching of two distributions. What's more, we found a growing peak near the 0 value. This is caused by the lack in trajectory information due to the growing time interval.

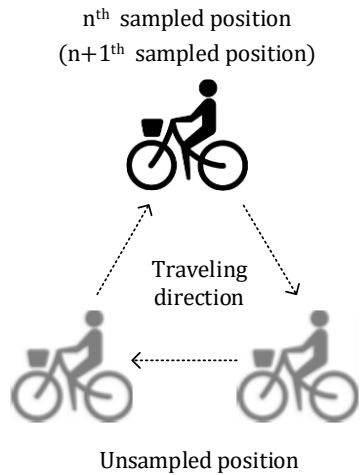

n<sup>th</sup> sampled position
(n+1<sup>th</sup> sampled position)

Traveling direction

Unsampled position

Figure 9. Zero velocity caused by long time interval

As shown in figure 9, suppose the traveler started from the position at top, passed by the position of two transparent and blurred icons and finally went back to origin or the position nearby. This is originally a round trip. However, the subsampling may only capture the origin and destination in this round trip so that the computed shortest distance between two sampled position is near 0 and the computed velocity is also nearly 0.

Consequently, in the trajectory with long time interval, it's very hard to distinguish the walking and bike travels by pure velocity. Therefore, in the training of detection model, we carefully choose the features to finish the job.

**Training of detection model**

By referencing to result of comparison and former studies that focus on the travel mode detection[8, 10, 11], we adopted the random forest method for the detection of bike travels in the study area. An important key step is the construction of features. In [8], author stated that velocity-related features can mostly help extract the walking trip and bike trip. In [9], besides velocity-related features, some global features like total travel distance and travel time can account for as large as the contribution of distinguishment of travel mode. Therefore, considering the special case that the time interval in our dataset is at least 5 minutes, we added some new features and finally adopted the ones listed in Table 2.

Table 2. The features of random forest model in our study:

| Feature Name | Physical Significance |
|---|---|
| 1. Distance | The total traveled distance in the moving segment |
| 2. Time | The time span of moving segment |
| 3. Points | The quantity of GPS records in the moving segment |
| 4. VCR | The average change rate of velocity |
| 5. MVCR | Maximal change rate of velocity |
| 6. MaxAcceleration | Maximal acceleration |
| 7. Avgspeed_1 | The timely average velocity |
| 8. Minspeed | Minimal velocity |
| 9. Maxspeed | Maximal velocity |
| 10. Avgspeed_2 | The quantitively average velocity |

Then, we will explain how these features are computed.

Since the GPS records are a series of tuple of position and time, the series can be implemented as:

$$GPS = [(P_1, t_1), (P_2, t_2), \ldots\ldots, (P_n, t_n)] \qquad (1)$$

Thus, the distance between the neighboring records should also be a series, the element of which is

$$D_i = Dis(P_i, P_{i+1}), i = 1, 2, \ldots\ldots, n-1 \qquad (2)$$

Where, the function *Dis()* is used to compute the shortest distance.

The feature 1 Distance equals to the summary of $D_i$:

$$Distance = sum(\{D_i | i = 1, 2, \ldots\ldots, n-1\}) \qquad (3)$$

Feature 2 Time equals to:

$$Time = t_n - t_1 \qquad (4)$$

The velocity can be computed by:

$$V_i = \frac{D_i}{t_{i+1} - t_i} \qquad (5)$$

Feature 8 Minspeed and feature 9 Maxspeed are separately the minimum and maximum of group of $V_i$.

In this study, we define the change rate of velocity as:

$$vcr_i = \frac{|V_{i+1} - V_i|}{V_i}, i = 1, 2, \ldots\ldots, n-2 \qquad (6)$$

Therefore, feature 4 VCR, feature 5 MVCR refers to the average and maximum of group of $vcr_i$ in one moving segment. Also, the acceleration is computed by:

$$acc_i = \frac{|V_{i+1} - V_i|}{t_{i+1} - t_i}, i = 1, 2, \ldots\ldots, n-2 \qquad (7)$$

Feature 6 MaxAcceleration is the maximum of group of $acc_i$; feature 7 Avgspeed_1 is:

$$Avgspeed\_1 = \frac{Distance}{Time} \qquad (8)$$

Feature 10 Avgspeed_2 is:

$$Avgspeed\_2 = \frac{sum(\{V_i | i=1,2,\ldots\ldots,n-2\})}{Points} \qquad (9)$$

The ground truth trips will be preprocessed by above equations and fed into the random forest model for training.

We computed the features in the subsampled trajectory set by using equation 1 to 9. Then, the subsampled trajectory set will be converted and aggregated as the dataset for training. The whole dataset is randomly divided to training set and test set by 80% and 20% to prevent over fitting. To furtherly test the reliability of model, we operated 10-fold cross validation on the model. The accuracy ranges from 89.29% to 100%. We adopted the model with test score of 100% to detect the bike and walking travels in the NTT DOCOMO dataset for big picture analysis.